%% file: paper.tex
\documentclass[runningheads]{llncs}
\usepackage[T1]{fontenc}
\usepackage{graphicx}
\usepackage{amsmath}
\usepackage{algorithm}
\usepackage{algorithmic}
\usepackage{hyperref}
\usepackage{cite}
\usepackage{amssymb}
\begin{document}
\title{RACER: Reinforced Agent Collaboration for Explainable Reasoning on Knowledge Graphs}
%
%\titlerunning{Abbreviated paper title}
% If the paper title is too long for the running head, you can set
% an abbreviated paper title here
%
\author{Yuwei Lou\inst{1}\orcidID{0009-0002-5058-6401} \and
Hao Hu \inst{1}\orcidID{0009-0001-8277-9876} \and
Yuzhou Jiang\inst{1}\orcidID{0009-0006-5881-8843} \and
Zongfei Zhang\inst{2}\orcidID{0009-0000-6702-8807} \and
Liang Wang \inst{1} \orcidID{0000-0001-5444-748X} \and
Jincai Liu \inst{3} \orcidID{0009-0003-4411-9847} \and
Jidong Ge \inst{1} \orcidID{0000-0003-1773-0942} \and
Xianping Tao \inst{1}\orcidID{0000-0002-5536-3891}} 

\authorrunning{Y. Lou et al.}
% First names are abbreviated in the running head.
% If there are more than two authors, 'et al.' is used.
%
\institute{State Key Laboratory for Novel Software Technology, Nanjing University, China
\email{\{yuweilou,jiangyuzhou\}@smail.nju.edu.cn, \{myou,wl,gjd,txp\}@nju.edu.cn}\\
Independent Researcher
\email{zhangzongfei007@gmail.com}\\
Chinaunicom Software Nanjing Branch
\email{liujc50@chinaunicom.cn}
}
\maketitle              % typeset the header of the contribution
\begin{abstract}
Large Language Models (LLMs) often suffer from hallucination and struggle with complex reasoning tasks requiring multi-hop domain knowledge. While integrating Knowledge Graphs (KGs) provides a structured and verifiable information source, current KG-enhanced LLM paradigms usually rely on single-agent path extraction and fixed prompting, lacking adaptability and facing huge search spaces. To address these challenges, we propose RACER, a Reinforced Agent Collaboration framework for Explainable Reasoning on knowledge graphs. RACER employs a semantic-aware action pruning and teacher-guided reinforcement learning mechanism to efficiently extract high-quality reasoning pathways from large-scale KGs. Furthermore, to mitigate single-path generation pitfalls, we introduce a cross-task accumulated shared memory graph paired with an attention-driven multi-path knowledge refinement module. Finally, RACER orchestrates these components through a four-role multi-agent collaboration system (GraphAgent, TemplateAgent, AnswerAgent, and CriticAgent) to dynamically refine prompts and evaluate answers. Extensive experiments on CommonsenseQA and OpenBookQA datasets demonstrate that RACER significantly outperforms state-of-the-art KG-enhanced LLM baselines with an average improvement of 5\%, offering robust and highly interpretable reasoning capabilities.
\end{abstract}

\input{chapter/introduction}
\input{chapter/relatedwork}

\input{chapter/model}

\input{chapter/experiments}
\input{chapter/conclusion}

\bibliographystyle{splncs04}
\bibliography{paper}

\end{document}

%% file: chapter/introduction.tex
\section{Introduction}
% 随着大语言模型（LLM）性能不断增加[0, 1]，一些以模型如Gemini3，Claude4.5的的代理开始在真实世界的广泛应用~\cite{dong2024clrbench,kung2023performance,minaee2024large,pan2024unifying}。尽管基于这些模型的代理表现的十分优秀，但是其基础架构决定的幻觉问题~\cite{dong2024modality}不可避免出现影响任务的准确性。尤其是大模型面临的问题是在自己训练的过程中未曾遇到或者是关联样本间距很长的情况下，往往效果很差~\cite{minaee2024large,pan2024unifying}。
As Large Language Models (LLMs) continue to demonstrate increasingly powerful capabilities~\cite{openai2023gpt4,anthropic2024claude3}, agents built upon models such as Gemini 3 and Claude 4.5 have begun to see widespread deployment in real-world applications ~\cite{dong2024clrbench,kung2023performance,minaee2024large,pan2024unifying}. Despite the impressive performance exhibited by these model-based agents, hallucination problems ~\cite{dong2024modality}, inherent to their underlying architecture, inevitably emerge and compromise task accuracy. A particularly critical challenge arises when LLMs encounter scenarios absent from their training data or involving long-range dependencies between related samples, where their effectiveness often degrades significantly~\cite{minaee2024large,pan2024unifying}.

% 检索增强生成（RAG）被视为缓解幻觉的速效方案。RAG可以利用外部问题相关的知识库来增加LLM运行的效率~\cite{lewis2020retrieval,minaee2024large,li2022survey},这样可以极大程度的降低LLM出现幻觉的情况，从而提升任务的准确率。但是由于提供的数据质量不同，或者是外部知识库对于问题相关的知识分布过散~\cite{li2022survey}，或者是需要结合多个外部知识才可以推理出正确知识，这些情况下导致运行RAG时无法检索到正确和问题匹配的知识出来。这些真实存在的情况导致RAG运行还是存在大量问题~\cite{pan2024unifying}。虽然有微软的GraphRAG~\cite{edge2024from}出现，利用知识图结构把“RAG”从“稠密向量检索”的性能往前推进了不少，但是核心检索步骤中知识图结构本身没有参与，也没有可学习的检索策略。
Retrieval-Augmented Generation (RAG) has been regarded as a promising solution for mitigating hallucinations. By leveraging external knowledge bases relevant to the query, RAG enhances the operational efficacy of LLMs ~\cite{lewis2020retrieval,minaee2024large,li2022survey}, substantially reducing hallucination occurrences and thereby improving task accuracy. However, due to variations in data quality, scattered distribution of query-relevant knowledge within external corpora ~\cite{li2022survey}, or the necessity of synthesizing multiple external knowledge sources for correct reasoning, RAG often fails to retrieve knowledge that accurately matches the query. These practical challenges significantly limit the effectiveness of RAG in real-world deployments ~\cite{pan2024unifying}. Although Microsoft's GraphRAG~\cite{edge2024from} advances RAG beyond dense vector retrieval by incorporating knowledge graph structures, the core retrieval process itself does not involve the graph structure in a learnable manner, nor does it incorporate trainable retrieval strategies.

% 结构化知识图谱（KG）~\cite{pan2024unifying,hu2023survey}以（头，关系，尾）三元组形式存储原子事实，天然支持多跳、可验证与即插即用更新~\cite{pan2024unifying,hu2023survey}。知识图谱中所存储的知识可以详细表示从一个建模节点到另一个建模节点的关系逻辑链路~\cite{hu2023survey,yang2024give,pan2024unifying}，这对于LLM执行复杂任务提供给了一种新的解决思路。目前主要的研究聚焦于基于知识图谱的固定知识用在大模型预训练的过程中去~\cite{sun2019ernie,peters2019knowledge}，另外一种研究主要关注如何利用知识图谱中的具有结构性逻辑性知识存储来增加大模型内部推理的过程，例如K-BERT~\cite{liu2020kbert}等。但是当前很多的顶尖大模型都是封闭不开源的，比如GPT，Gemini，Claude等系列。于是当前的很多研究都聚焦于如何用知识图谱中的结构化知识来构建大模型提示词工程~\cite{liu2023pretrain,pan2024unifying}，比如Cok~\cite{wang2023cok}将llm生成的推理逻辑和外部的知识图谱知识形成对比来论证真实性，KGR~\cite{guan2024mitigating}利用知识图谱逐步提取精炼知识，从而优化llm思维过程，KnowGPT~\cite{zhang2024knowgpt}提出了一种在知识图谱中利用机器学习提取推理链路来，配合模板来增加llm性能。
Structured Knowledge Graphs (KGs)~\cite{pan2024unifying,hu2023survey} store atomic facts as (head, relation, tail) triples, inherently supporting multi-hop reasoning, verifiability, and plug-and-play updates ~\cite{pan2024unifying,hu2023survey,yang2024give,pan2024unifying}. The knowledge encoded in KGs explicitly represents relational logical pathways from one entity to another ~\cite{hu2023survey,yang2024give,pan2024unifying}, offering a novel paradigm for LLMs to execute complex tasks. Current research primarily focuses on incorporating fixed KG knowledge during LLM pre-training ~\cite{sun2019ernie,peters2019knowledge}, while another line of work investigates leveraging structural and logical knowledge within KGs to augment internal LLM reasoning processes, as exemplified by K-BERT ~\cite{liu2020kbert}. However, many state-of-the-art LLMs, such as GPT, Gemini, and Claude series, remain closed-source and inaccessible. Consequently, recent efforts have shifted toward utilizing structured KG knowledge for prompt engineering with LLMs ~\cite{liu2023pretrain,pan2024unifying}. For instance, CoK ~\cite{wang2023cok} validates the faithfulness of LLM-generated reasoning chains by contrasting them with external KG knowledge; KGR ~\cite{guan2024mitigating} progressively extracts and refines knowledge from KGs to optimize LLM reasoning trajectories; and KnowGPT~\cite{zhang2024knowgpt} proposes extracting reasoning paths from KGs through machine learning techniques, combined with templates, to enhance LLM performance.

% 虽然当前知识图谱增加大模型领域的研究进展良好，但是任然面临以下几个问题：1. 搜索空间的巨大。这导致目前目前检索困难，即使有研究~\cite{zhang2024knowgpt,wen2023mindmap}在使用机器学习的手段去完成，但是效率不高。2. 固定的模板限制，现在大部分给LLM执行的模板都是利用专家提前设计，哪怕是KnowGPT~\cite{zhang2024knowgpt}中都是模板大部分提前设计并且利用多臂赌博机在任务中动态选择，实测下来效果不佳。3. 任务执行效率提升局限，通过知识图谱增加提示词工程是一个绝佳的想法，但是大部分直接交付单个大模型去执行，这完全忽视了目前MAS研究领域的优势。
Despite the promising progress in integrating Knowledge Graphs with LLMs, several critical challenges remain: (1) Enormous search space. The vast scale of KGs renders retrieval inherently difficult. Although some approaches~\cite{zhang2024knowgpt,wen2023mindmap} employ machine learning techniques to address this, their efficiency remains limited. (2) Fixed template constraints. Most templates designed for LLM execution rely on handcrafted expert rules. Even in KnowGPT, templates are predominantly pre-designed and dynamically selected via multi-armed bandits during task execution, which demonstrates suboptimal performance in practice. (3) Limited task execution efficiency. While enhancing prompt engineering through KGs presents a compelling direction, most existing approaches delegate tasks to a single LLM, completely overlooking the advantages of Multi-Agent Systems (MAS) research.

% 为此，我们提出了RACER: Reinforced Agent Collaboration for Explainable Reasoning on Knowledge Graphs。一个利用强化学习，知识图谱和多智能体协同的任务增强型框架。它利用强化学习提取知识图谱中的关键知识链路，并且经过知识精炼与问题强相关，同时利用多智能体协同的步骤让链路更好的服务于任务执行，极大提升了任务执行效率。具体贡献如下：
  % 1. 我们提出了一种结合语义感知动作剪枝和教师引导的强化学习（RL）框架，用于在大规模知识图谱上针对问题输入提取高效的推理链路。
  % 2. 我们设计了一个跨任务累积的共享记忆图，记录每条边的全局统计信息，并将其融入路径选择和精炼过程中。通过双重注意力机制对多条候选路径进行精炼，生成高质量的自然语言知识片段，为下游语言模型提供可解释的知识支持。
  % 3. 我们基于上述的贡献构建了一个包含四个角色的多智能体协作推理框架,通过分工明确的协同机制，实现了从路径搜索到知识生成再到最终问答的全流程优化。
  % 4. 我们在多个基准数据集上系统性验证了所提方法的有效性，相比于最新的方法性能均有5%左右提升。
To address these challenges, we propose \textbf{RACER: Reinforced Agent Collaboration for Explainable Reasoning on Knowledge Graphs}, a task-enhancement framework that integrates reinforcement learning, knowledge graphs, and multi-agent collaboration. RACER leverages reinforcement learning to extract critical reasoning paths from knowledge graphs, refining knowledge to establish strong relevance with the query. Furthermore, it employs multi-agent collaborative procedures to align these paths more effectively with task execution, substantially improving operational efficiency. Our key contributions are summarized as follows:
\begin{enumerate}
    \item We propose a \textbf{semantic-aware action pruning} and \textbf{teacher-guided reinforcement learning (RL)} framework for extracting efficient reasoning paths over large-scale knowledge graphs conditioned on query inputs.
    
    \item We design a \textbf{cross-task accumulated shared memory graph} that records global statistical information for each edge, integrating this knowledge into path selection and refinement processes. Through a \textbf{dual attention mechanism}, we refine multiple candidate paths to generate high-quality natural language knowledge snippets, providing interpretable knowledge support for downstream language models.
    
    \item We construct a \textbf{four-role multi-agent collaborative reasoning framework} built upon the aforementioned contributions. Through well-defined division of labor and active feedback mechanisms (including a closed-loop review by the CriticAgent), we achieve end-to-end optimization spanning from path search to knowledge generation and final question answering.
    
    \item We systematically validate the effectiveness of our proposed method across multiple benchmark datasets, achieving approximately \textbf{5\% performance improvement} over state-of-the-art baselines.
\end{enumerate}

%% file: chapter/relatedwork.tex
\section{Related Work}

To address the challenges in complex reasoning tasks, specifically hallucination and logical interpretability in LLMs, researchers have proposed various methods. We broadly categorize them into three main areas: Retrieval-Augmented Generation for LLMs, Integration of Knowledge Graphs (KGs) and LLMs, and Multi-Agent Collaboration Frameworks.

\subsection{Retrieually-Augmented Generation for LLMs}

Recently, Retrieval-Augmented Generation (RAG) models have been extensively explored to enhance LLMs with external knowledge from text corpora or online sources~\cite{lewis2020retrieval, minaee2024large, li2022survey}. Combining LLMs with external knowledge retrieval systems can substantially reduce hallucination. However, these approaches face profound challenges in domain-specific or complex applications: (i) Data quality and scattered distribution. Domain knowledge is often scattered across diverse and unstructured sources~\cite{li2022survey}, leading to potential inconsistencies or errors in the retrieved knowledge~\cite{pan2024unifying}. (ii) Knowledge hierarchy and complex searching. Textual documents typically lack explicit relationships and structured organization, limiting the reasoning and inference capabilities of basic RAG models. Furthermore, finding relevant terminology in huge search spaces can be computationally expensive and time-consuming.

\subsection{Integration of KGs and LLMs}

KG-enhanced LLMs mitigate these unstructrued retrieval issues by leveraging the structured relational knowledge in Knowledge Graphs (KGs) to ground the model's responses in established facts and verifiably logical pathways~\cite{hu2023survey, yang2024give, pan2024unifying}.

\paragraph{Integrating KGs during Training.} Earlier studies adopted heuristic methods to inject knowledge from KGs into the LLMs during pre-training or fine-tuning, such as ERNIE~\cite{sun2019ernie} and KnowBERT~\cite{peters2019knowledge}, which incorporate entity embeddings and align them with word representations.

\paragraph{KG Prompting for LLMs.} Since most state-of-the-art LLMs (e.g., GPT-4~\cite{openai2023gpt4}, Claude~\cite{anthropic2024claude3}) are confined to a black-box role via APIs, research focus has recently shifted towards KG prompting that enhances fixed LLMs with structured prompts. For instance, CoK~\cite{wang2023cok} introduces a Chain-of-Knowledge prompting to decompose LLM-generated reasoning chains into evidence triples, verifying them natively against external KGs. KGR~\cite{guan2024mitigating} uses KGs to retroactively verify the initial draft responses generated by LLMs. More recently, KnowGPT~\cite{zhang2024knowgpt} leverages multi-armed bandits to extract paths from KGs and applies context-aware modules for prompt construction. 

While KG prompting is promising, the reliance on fixed, manually designed prompt templates restricts flexibility across diverse semantic contexts. Existing KG integration heavily relies on single-LLM executions, completely ignoring the structural advantages of Multi-Agent Systems.

\subsection{Multi-Agent Collaboration Frameworks}

To overcome the performance limits of single-query LLMs on complex knowledge reasoning, recent studies have begun employing multi-agent frameworks~\cite{wu2023autogen, hong2023metagpt, qian2023chatdev}, where different agents handle distinct reasoning, validation, and generation modules. Multi-agent systems (MAS) divide sophisticated tasks into cooperative sub-roles, mitigating single-agent failure points and improving overall reliability. For instance, AutoGen~\cite{wu2023autogen} provides a generalized framework for conversable agents, while frameworks like MetaGPT~\cite{hong2023metagpt} and ChatDev~\cite{qian2023chatdev} assign specialized architectural roles to agents for collaborative problem-solving. In contrast to traditional single-agent KG reasoning models, our proposed RACER explicitly incorporates Reinforced Agent Collaboration to orchestrate path searching, knowledge refinement, and answer generation in a synergized pipeline.

%% file: chapter/model.tex
\section{Methodology}

\begin{figure}[htbp]
    \centering
    \includegraphics[width=\textwidth]{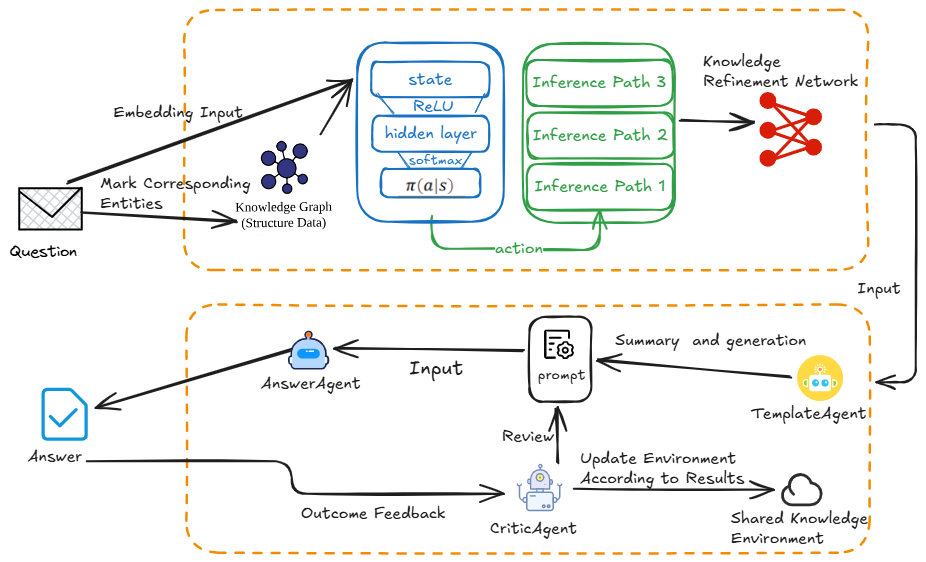}
    \caption{The overall architecture of the proposed RACER framework. It illustrates the semantic-aware reinforcement learning path searching, the shared memory graph refinement process, and the multi-agent collaboration workflow.}
    \label{fig:framework}
\end{figure}

In this section, we introduce the RACER framework, a reinforced multi-agent collaborative system for explainable reasoning on knowledge graphs. As illustrated in Figure~\ref{fig:framework}, RACER consists of three main components: (1) a Semantic-Pruning Knowledge Graph Reinforcement Learning module that extracts critical reasoning paths; (2) a Shared Memory Graph with Dual Attention that records global path statistics and refines candidate paths into high-quality natural language knowledge; and (3) a Multi-Agent Collaborative Reasoning Framework that orchestrates specialized agents (GraphAgent, TemplateAgent, AnswerAgent, and CriticAgent) to seamlessly execute the entire reasoning and answering pipeline.

\subsection{Semantic-Pruning Knowledge Graph Reinforcement Learning}
% 语义感知动作剪枝和教师引导的强化学习
% KnowGPT~\cite{zhang2024knowgpt}中指出理想子图GSub需要包含尽可能多的知识图谱中相关的源实体Qs和目标实体Qt，同时保持与问题上下文的强相关性，并且要简洁无冗余，以便能够以有限长度输入到LLM中。由于从大规模知识图谱中提取子图是NP难问题，找到满足上述条件的GSub图非常困难。我们认为在理想子图GSub中从源实体Qs和目标实体Qt的路径是由多条和问题相关的逻辑推理链路组成的。那么说构建理想子图GSub的另外一种等价方法就是最大可能找到每一条这样的推理链路。为了高效找到这样的推理链路，我们提出了一种基于语义感知的动作空间剪枝和教师引导的强化学习方法。通过预训练语言模型（PLM）嵌入统一表示问题、关系和节点，从而实现高效的语义匹配。
KnowGPT~\cite{zhang2024knowgpt} identifies that an ideal subgraph $\mathcal{G}_{\text{sub}}$ should contain as many relevant source entities $\mathcal{Q}_s$ and target entities $\mathcal{Q}_t$ from the knowledge graph as possible, while maintaining strong relevance to the query context and remaining concise without redundancy, thereby enabling input to LLMs within limited context lengths. However, extracting such subgraphs from large-scale knowledge graphs is NP-hard, making the identification of $\mathcal{G}_{\text{sub}}$ satisfying all aforementioned criteria computationally intractable.

We posit that paths connecting source entities $\mathcal{Q}_s$ and target entities $\mathcal{Q}_t$ within the ideal subgraph $\mathcal{G}_{\text{sub}}$ consist of multiple logical reasoning chains relevant to the query. Consequently, constructing $\mathcal{G}_{\text{sub}}$ is equivalently formulated as maximizing the identification of such reasoning chains. To efficiently discover these paths, we propose a \textbf{semantic-aware action space pruning} and \textbf{teacher-guided reinforcement learning} approach. Specifically, we employ Pre-trained Language Models (PLMs) to embed queries, relations, and entities into a unified representation space, thereby enabling efficient semantic matching.

\subsubsection{Semantic Action Space Pruning}
% 语义剪枝的核心思想是通过计算候选动作与问题的语义相似度，动态压缩动作空间，减少不必要的搜索分支。
The core idea of semantic pruning is to dynamically compress the action space by computing semantic similarity between candidate actions and the query, thereby eliminating unnecessary search branches.
% 动作评分:对于当前节点vt的候选动作集合A我们计算每个动作的综合得分，如下所示 
For the candidate action set $ A = \{(n_i, r_i)\}_{i=1}^M $  of the current node $v_t$, we compute the composite score for each action as follows:
\begin{equation}
\text{score}(n_i, r_i) = \alpha \cdot \text{sim}(r_i, \mathbf{q}) + \beta \cdot \text{sim}(n_i, \mathbf{q}) + \gamma \cdot \text{sim}(n_i, \mathbf{g})
\end{equation}

% 其中：q  和 g  分别为问题和目标节点的嵌入表示，sim(⋅)  为余弦相似度，用于衡量语义相关性。α+β+γ=1 ，且 α,β,γ  是超参数，用于平衡不同维度的相似度。
where $\mathbf{q}$ and $\mathbf{g}$ denote the dense embedding representations of the query and the target node, respectively, and $\text{sim}(\cdot)$ denotes cosine similarity for measuring semantic relevance. The hyperparameters satisfy $\alpha + \beta + \gamma = 1$, where $\alpha, \beta, \gamma$ balance different similarity dimensions.

% 动作选择:根据上述评分公式，我们选择得分最高的前 K  个动作作为候选动作，显著减少搜索空间。这一过程不仅提高了搜索效率，还确保了候选动作与问题语义对齐。
Based on the scoring function above, we select the top-$K$ actions with the highest scores as the candidate set, significantly reducing the search space. This procedure not only improves search efficiency but also ensures semantic alignment between candidate actions and the query.

\subsubsection{Teacher-Guided Policy Gradient Training}
% 为了进一步提升策略的泛化能力，我们引入了基于图最短路径的教师引导机制。教师轨迹通过 BFS 算法计算得到，为策略训练提供了强先验。
To further enhance the generalization capability of the policy, we introduce a graph shortest path-based teacher guidance mechanism. Teacher trajectories are computed via the BFS algorithm, providing strong priors for policy training.

% 教师轨迹计算:对于每个问题，我们使用 BFS 算法计算从起始节点到目标节点的最短路径 T 。教师轨迹 T 提供了最优路径的先验信息。
For each query, we employ the BFS algorithm to compute the shortest path $ P^* = [v_0, v_1, \dots, v_T] $ from the source node $v_0$ to the target node $v_T$. The teacher trajectory $P^*$ provides prior information regarding the optimal path.

% 混合训练机制: 在训练过程中，我们以高概率 p teacher遵循 Teacher 轨迹中的下一步动作，其余时间由策略网络采样。具体公式如下：
During training, we follow the next action from the teacher trajectory with high probability $p_{\text{teacher}}$, while sampling from the policy network for the remaining steps. The specific formulation is as follows:

\begin{equation}
\pi(a \mid s_t) = \begin{cases} \mathcal{T}(a \mid s_t), & \text{with probability } p_{\text{teacher}} \\ \pi_{\text{RL}}(a \mid s_t), & \text{with probability } 1 - p_{\text{teacher}} \end{cases}
\end{equation}

% 其中，T(a∣s)  是 Teacher 轨迹提供的动作概率，πRL(a∣s)  是基于强化学习的策略。
where $\mathcal{T}(a \mid s_t)$ denotes the action probability provided by the teacher trajectory, and $\pi_{\text{RL}}(a \mid s_t)$ represents the reinforcement learning-based policy. The state $s_t$ is mathematically formulated as $s_t = [\mathbf{v}_t; \mathbf{q}; \mathbf{h}_t]$, integrating the current node embedding, the query embedding, and the historical path encoding.
% Teacher 监督信号
% 对于 Teacher 指导的步骤，我们使用固定正回报 Rteacher=1.0  作为监督信号，加速策略收敛。这一机制不仅提供了强先验，还显著提高了策略的泛化能力。
\paragraph{Teacher Supervision Signal.}
For steps guided by the teacher, we employ a fixed positive reward $R_{\text{teacher}} = 1.0$ as the supervision signal to accelerate policy convergence. This mechanism not only provides strong priors but also significantly enhances the generalization capability of the policy.

\subsubsection{Reward Function Design}
% 为了使策略在语义空间中单调逼近目标实体，同时提供足够的探索激励，我们设计了一个综合的奖励函数，包含多个组成部分，每个部分都有其特定的作用。奖励函数的设计旨在平衡探索效率和语义相关性，确保策略能够高效地找到目标节点。
To enable the policy to monotonically approach the target entity in semantic space while providing sufficient exploration incentives, we design a composite reward function comprising multiple components, each serving a specific purpose. The reward formulation aims to balance exploration efficiency and semantic relevance, ensuring effective target node identification.

% 基础步惩罚（Step Penalty）：为了惩罚不必要的探索步骤，我们在每一步中引入一个基础步惩罚项 r_{step} 该惩罚项是一个负常数，用于减少路径长度，避免策略陷入冗长的探索过程：
To penalize unnecessary exploration steps, we introduce a base step penalty $r_{\text{step}}$ at each timestep. This penalty term is a negative constant designed to reduce path length and prevent the policy from engaging in protracted exploration:

% 关系相似度奖励:为了鼓励策略选择与问题语义相关的边，我们引入了一个关系相似度奖励项 r_{\text{rel}} = \lambda_{\text{rel}} \cdot \text{sim}(r_t, q) 。该奖励项基于当前边与问题的语义相似度计算.
\paragraph{Relation Similarity Reward.}
To encourage the policy to select edges semantically relevant to the query, we introduce a relation similarity reward term $r_{\text{rel}} = \lambda_{\text{rel}} \cdot \text{sim}(r_t, \mathbf{q})$, where $\lambda_{\text{rel}}$ is a scaling coefficient. This reward is computed based on the semantic similarity between the traversed relation $r_t$ and the query embedding $\mathbf{q}$.

% 目标密集奖励（Goal-Oriented Dense Reward）: 为了引导策略单调逼近目标节点，我们设计了一个目标密集奖励项 rgoal.该奖励项基于当前节点与目标节点的语义相似度变化计算：
\paragraph{Goal-Oriented Dense Reward.}
To guide the policy to monotonically approach the target node, we design a goal-oriented dense reward term $r_{\text{goal}}$. This reward is computed based on the change in semantic similarity between the current node and the target node: 
\begin{equation}
r_{\text{goal}} = \lambda_{\text{goal}} \cdot (\text{sim}(v_{t+1}, \mathbf{g}) - \text{sim}(v_t, \mathbf{g}))
\end{equation}
% 其中，\lambda_{\text{goal}} >0  是一个超参数，用于调整目标密集奖励的权重；sim(⋅)是余弦相似度函数，用于衡量当前节点v_t和目标节点g  的语义相似度。
where $\lambda_{\text{goal}} > 0$ is a hyperparameter controlling the weight of the goal-oriented dense reward, and $\text{sim}(\cdot)$ denotes the cosine similarity function measuring semantic similarity between the embeddings of current node $v_t$ and the target node $\mathbf{g}$. During the inference phase for multiple-choice QA, since the optimal target is unknown, the target representation $\mathbf{g}$ is instantiated by mapping each candidate answer choice to its corresponding graph entity representation, allowing RACER to independently evaluate paths leading towards each candidate option.

% 到达目标奖励（Reach Goal Reward）为了激励策略快速到达目标节点，我们引入了一个到达目标奖励项 r_reach当策略成功到达目标节点时，给予一个较大的正奖励：
\paragraph{Reach Goal Reward.}
To incentivize the policy to reach the target node efficiently, we introduce a reach goal reward term $r_{\text{reach}}$. When the policy successfully arrives at the target node $v_T$, a substantial positive reward is granted:
\begin{equation}
    r_{\text{reach}} = \begin{cases} \lambda_{\text{reach}}, & \text{if } v_{t+1} = v_T \\ 0, & \text{otherwise} \end{cases}
\end{equation}

% 其中，λ_reach>0是一个超参数，用于控制到达目标奖励的强度。
where $\lambda_{\text{reach}} > 0$ is a hyperparameter governing the magnitude of the reach goal reward.

% 总奖励函数（Total Reward Function）:综合上述各项奖励，总奖励函数 R 定义为：
\paragraph{Total Reward Function.}
Integrating the aforementioned reward components, the total reward function $\mathcal{R}$ is defined as:
\begin{equation}
\mathcal{R} = r_{\text{step}} + r_{\text{rel}} + r_{\text{goal}} + r_{\text{reach}}
\end{equation}
% 这种奖励函数设计不仅提供了密集的奖励信号，还确保策略在语义空间中单调逼近目标实体，显著提高了推理的准确性和效率。通过调整超参数 λ_step、λ_rel、λ _goal和 λ_reach，可以灵活地平衡不同奖励项的权重，以适应不同的任务需求。
This reward design not only provides dense reward signals but also ensures that the policy monotonically approaches the target entity in semantic space, significantly improving both reasoning accuracy and efficiency. By adjusting the hyperparameters $\lambda_{\text{step}}$, $\lambda_{\text{rel}}$, $\lambda_{\text{goal}}$, and $\lambda_{\text{reach}}$, the relative weights of different reward terms can be flexibly balanced to accommodate diverse task requirements.

\subsection{Shared Memory Graph and Attention-Driven Knowledge Refinement}
% 传统知识图谱推理方法在生成路径时缺乏对历史路径质量的有效记忆和利用。此外，单路径搜索容易陷入局部最优，难以提供高质量的辅助知识，限制了下游语言模型（LLM）的性能。为了克服这些挑战，我们提出了一种结合共享记忆图和注意力驱动的多路径知识精炼方法，通过持久化边级统计和智能路径选择，持续提升知识路径的质量。
Traditional knowledge graph reasoning methods lack effective memory and utilization of historical path quality during path generation. Moreover, single-path search is prone to local optima, failing to provide high-quality auxiliary knowledge and thereby limiting the performance of downstream Large Language Models (LLMs). To address these limitations, we propose a shared memory graph coupled with attention-driven multi-path knowledge refinement method. This approach persistently maintains edge-level statistics and enables intelligent path selection, continuously enhancing the quality of knowledge paths.
\subsubsection{Shared Memory Graph}
% 共享记忆图的核心思想是通过记录每条边的历史表现，为路径选择提供全局统计信息，从而优化路径质量。
The core idea of the shared memory graph is to record the historical performance of each edge, providing global statistical information for path selection to optimize path quality.
% 对于知识图谱中的每条边 e=(h,r,t) ，我们维护以下三元统计信息：
% 其中：succ(e)表示该边出现在成功路径中的次数。fail(e)表示该边出现在失败路径中的次数。reject(e)表示该边被拒绝的次数。

For each edge $e = (h, r, t)$ in the knowledge graph, we maintain the following triplet of statistics:
\begin{equation}
\text{stat}(e) = \{\text{succ}(e), \text{fail}(e), \text{reject}(e)\}
\end{equation}
where $\text{succ}(e)$ denotes the number of times edge $e$ appears in successful paths, $\text{fail}(e)$ denotes the number of times it appears in failed paths, and $\text{reject}(e)$ denotes the number of times it has been rejected during the pruning process.

% 边质量得分计算：基于上述统计信息，我们计算每条边的质量得分，用于指导路径选择：% 其中，γ  是一个超参数，用于平衡成功路径与拒绝路径的影响。
\paragraph{Edge Quality Scoring.}
Based on the aforementioned statistics, we compute a quality score for each edge to guide path selection:
\begin{equation}
\text{score}(e) = \frac{\text{succ}(e)+1}{\text{freq}(e)+2} - \gamma \cdot \frac{\text{reject}(e)}{\text{total}(e)}
\end{equation}
where $\text{freq}(e)=\text{succ}(e)+\text{fail}(e)$ and $\text{total}(e)=\text{freq}(e)+\text{reject}(e)$. The hyperparameter $\gamma$ balances success against rejection rates.

% 动作选择优化：在动作选择时，根据边质量得分对候选动作进行重排序，优先选择历史表现好的边。这一机制不仅利用了全局统计信息，还避免了局部最优解。
\paragraph{Action Selection Optimization.}
During action selection, candidate actions are re-ranked according to edge quality scores, prioritizing edges with superior historical performance. This mechanism not only leverages global statistical information but also mitigates local optima.

\subsubsection{Multi-Path Generation}
% 为了生成多样化的高质量路径，我们设计了一种多路径生成机制，结合温度采样和贪心策略，平衡路径的多样性和质量。
To generate diverse high-quality paths, we design a multi-path generation mechanism that combines temperature sampling and greedy strategies to balance path diversity and quality.

% 路径生成过程：对于给定问题q和起点集合S，生成N条候选路径P = \{P_1, P_2, \dots, P_N\}。路径生成过程如下：
Given a query $\mathbf{q}$ and a starting node set $\mathcal{S}$, we generate $N$ candidate paths $\mathcal{P} = \{P_1, P_2, \dots, P_N\}$. The path generation process is formulated as:
\begin{equation}
a_t = 
\begin{cases} 
\text{sample}\big(\pi_\theta(\cdot \mid s_t), \tau\big), & \text{if } t < \frac{N}{2}, \\[6pt]
\arg\max_a \pi_\theta(a \mid s_t), & \text{otherwise},
\end{cases}
\end{equation}
% 其中，\pi_\theta(\cdot|s_t)是策略网络在状态st下的动作概率分布，\tau是温度参数，用于控制采样的随机性。在路径生成的前半部分使用温度采样增加多样性，后半部分使用贪心策略保证质量。
where $\pi_\theta(\cdot \mid s_t)$ denotes the action probability distribution of the policy network at state $s_t$, and $\tau$ is the temperature parameter controlling sampling randomness. Temperature sampling is employed during the first half of path generation to enhance diversity, while the greedy strategy is adopted in the second half to ensure quality.

\subsubsection{Attention-Driven Path Refinement}
% 为了从多条候选路径中提取关键信息，我们设计了一种轻量级的双重注意力路径精炼网络，通过跨路径关系建模与问题语义对齐，筛选出最相关的路径子集。

% \paragraph{问题定义.} 给定问题 $q$ 与 RL 生成的 $k$ 条候选路径 $\mathcal{P} = \{P_1, \dots, P_k\}$，学习精炼函数 $\mathcal{R}: (q, \mathcal{P}) \to \mathcal{P}^*$，其中 $\mathcal{P}^* \subseteq \mathcal{P}$ 且 $|\mathcal{P}^*| \ll k$，使得精炼后的路径集合保留回答问题所需的最关键信息。

% \paragraph{双重注意力架构.} 采用预训练语言模型对路径和问题进行编码：
% \[
% \mathbf{p}_i = \mathbf{W}_p \cdot \text{PLM}(P_i), \quad \mathbf{q} = \mathbf{W}_q \cdot \text{PLM}(q)
% \]
% 其中 $\mathbf{W}_p, \mathbf{W}_q \in \mathbb{R}^{d \times d_{\text{PLM}}}$ 为可学习的投影矩阵。

% \paragraph{跨路径自注意力.} 为建模路径间的互补与冗余关系，引入 $L$ 层多头自注意力：
% \[
% \mathbf{H}^{(l)} = \text{MultiHead-SelfAttn}(\mathbf{H}^{(l-1)}), \quad \mathbf{H}^{(0)} = [\mathbf{p}_1; \dots; \mathbf{p}_k]
% \]
% 输出精炼后的路径表示 $\mathbf{H}^{(L)} \in \mathbb{R}^{k \times d}$。

% \paragraph{问题-路径交叉注意力.} 计算问题对每条路径的语义匹配度：
% \[
% \boldsymbol{\alpha} = \text{softmax}\left(\frac{(\mathbf{1}_k \otimes \mathbf{q}) (\mathbf{H}^{(L)})^\top}{\sqrt{d}}\right) \cdot \mathbf{1}_d \in \mathbb{R}^k
% \]

% \paragraph{重要性评分与路径选择.} 引入可学习的评分器评估路径内在重要性：
% \[
% \boldsymbol{\beta} = \sigma(\mathbf{W}_2 \cdot \text{ReLU}(\mathbf{W}_1 (\mathbf{H}^{(L)})^\top)) \in \mathbb{R}^k
% \]
% 综合得分 $s_i = \lambda \alpha_i + (1-\lambda)\beta_i$，选择 Top-$k'$ 路径：
% \[
% \mathcal{P}^* = \{P_i \mid i \in \text{Top-}k'(\{s_1, \dots, s_k\})\}
% \]
% 其中 $\lambda$ 为可学习参数，$k'$ 通过验证集选取。精炼后的路径集 $\mathcal{P}^*$ 送入模板生成器为 LLM 提供紧凑且高质的上下文。
To extract critical information from multiple candidate paths, we design a lightweight dual-attention path refinement network that filters the most relevant path subset through cross-path relational modeling and query semantic alignment.

\paragraph{Problem Formulation.}
Given a query $q$ and $k$ candidate paths $\mathcal{P} = \{P_1, \dots, P_k\}$ generated by RL, we learn a refinement function $\mathcal{R}: (q, \mathcal{P}) \to \mathcal{P}^*$, where $\mathcal{P}^* \subseteq \mathcal{P}$ and $|\mathcal{P}^*| \ll k$, such that the refined path set preserves the most critical information required for answering the query.

\paragraph{Dual-Attention Architecture.}
We employ pre-trained language models to encode paths and the query:
\begin{equation}
\mathbf{p}_i = \mathbf{W}_p \cdot \text{PLM}(P_i), \quad \mathbf{q} = \mathbf{W}_q \cdot \text{PLM}(q)
\end{equation}
where $\mathbf{W}_p, \mathbf{W}_q \in \mathbb{R}^{d \times d_{\text{PLM}}}$ are learnable projection matrices.

\paragraph{Cross-Path Self-Attention.}
To model complementary and redundant relationships among paths, we introduce $L$ layers of multi-head self-attention:
\begin{equation}
\mathbf{H}^{(l)} = \text{MultiHead-SelfAttn}(\mathbf{H}^{(l-1)}), \quad \mathbf{H}^{(0)} = [\mathbf{p}_1; \dots; \mathbf{p}_k]
\end{equation}
outputting refined path representations $\mathbf{H}^{(L)} \in \mathbb{R}^{k \times d}$.

\paragraph{Query-Path Cross-Attention.}
We compute semantic matching scores between the query and each path. The query vector is expanded to match the number of paths via a Kronecker product, and the attention weights are acquired:
\begin{equation}
\boldsymbol{\alpha} = \text{softmax}\left(\frac{(\mathbf{1}_k \otimes \mathbf{q}) (\mathbf{H}^{(L)})^\top}{\sqrt{d}}\right) \cdot \mathbf{1}_d \in \mathbb{R}^k
\end{equation}

\paragraph{Importance Scoring and Path Selection.}
We introduce a learnable scorer to assess the intrinsic importance of paths:
\begin{equation}
\boldsymbol{\beta} = \sigma(\mathbf{W}_2 \cdot \text{ReLU}(\mathbf{W}_1 (\mathbf{H}^{(L)})^\top)) \in \mathbb{R}^k
\end{equation}
The composite score is computed as $s_i = \lambda \alpha_i + (1-\lambda)\beta_i$, and the Top-$k'$ paths are selected:
\begin{equation}
\mathcal{P}^* = \{P_i \mid i \in \text{Top-}k'(\{s_1, \dots, s_k\})\}
\end{equation}
where $\lambda$ is a learnable parameter and $k'$ is determined via validation. The refined path set $\mathcal{P}^*$ is fed into the template generator to provide compact and high-quality context for the LLM.

\subsection{Multi-Agent Collaboration Framework}
\subsubsection{Agent Roles and Collaboration}
% 我们将前面的所提到的模块结合代理，构建了一个包含四个角色的多智能体协作框架，通过分工明确的协同机制，实现从路径搜索到知识生成再到最终问答的全流程优化。
% GraphAgent:GraphAgent具有在知识图谱上使用语义感知动作剪枝和教师引导的强化学习策略网络生成知识三元组序列的能力，同时其集成Attention-Driven Knowledge Refinement来精炼知识三元组，使其成为高质量知识。
% TemplateAgent:将GraphAgent推理出来的高质量知识三元组序列转化为知识逻辑路径，并且和共享记忆图合并生成为大模型所理解的自然语言知识段模板。
% AnswerAgent：接收TemplateAgent生成的模板，并且按照任务要求进行回答问题。同时自动记录生成的答案正确性，并且接受CriticAgent的检查。
% CriticAgent：检查整个问题推理回答过程和最终的结果，输出决策d ∈ { ACCEPT , UNSURE } 和置信度c ∈ [ 0 , 1 ]。同时根据置信度对于决策信息更新到共享知识图中去。
We integrate the aforementioned modules into a collaborative multi-agent system, constructing a four-role multi-agent framework that achieves end-to-end optimization from path search to knowledge generation and final question answering through well-defined division of labor and coordination mechanisms.

\paragraph{GraphAgent.}
GraphAgent is equipped with the capability to generate knowledge triplet sequences over knowledge graphs using semantic-aware action pruning and teacher-guided reinforcement learning policy networks. Additionally, it integrates Attention-Driven Knowledge Refinement to distill high-quality knowledge triplets.

\paragraph{TemplateAgent.}
TemplateAgent transforms the high-quality knowledge triplet sequences inferred by GraphAgent into logical knowledge paths, and merges them with the shared memory graph to generate natural language knowledge segment templates comprehensible to large language models.

\paragraph{AnswerAgent.}
AnswerAgent receives templates generated by TemplateAgent and produces answers according to task requirements. It automatically records the correctness of generated answers and undergoes inspection by CriticAgent.

\paragraph{CriticAgent.}
CriticAgent examines the entire reasoning and answering process as well as the final results, outputting a decision $d \in \{\text{ACCEPT}, \text{UNSURE}\}$ and a confidence score $c \in [0, 1]$. Furthermore, if the CriticAgent outputs an UNSURE decision, it triggers an active feedback loop, signaling the GraphAgent to explore alternative semantic branches. The failure trajectories are simultaneously penalized in the Shared Memory Graph to prevent redundant cycles, establishing a highly adaptive self-correction operational loop.

%% file: chapter/experiments.tex
\section{Experiment}
We evaluate RACER on two widely recognized commonsense reasoning datasets and conduct extensive experiments. Our experiments address the following research questions:

\textbf{RQ1 (Main results):} How much does RACER improve over recent KG-enhanced LLM methods?

\textbf{RQ2 (Ablation Study):} Which components contribute critically to RACER's performance?

\subsection{Experimental Setup}
% 我们选择 CommonsenseQA 和 OpenBookQA 作为常识推理基准数据集。CommonsenseQA 包含 12,102 个需要常识知识的多项选择题；OpenBookQA 包含 5,957 个需要基础科学事实的多项选择题。两个数据集均以 ConceptNet 作为背景知识图谱
\paragraph{Datasets.} We select CommonsenseQA~\cite{talmor2019commonsenseqa} and OpenBookQA~\cite{mihaylov2018openbookqa} as benchmarks. CommonsenseQA contains 12,102 multiple-choice questions requiring commonsense knowledge. OpenBookQA contains 5,957 questions requiring elementary science facts. Both use ConceptNet~\cite{speer2017conceptnet} as the background KG.

\paragraph{Baselines.}
We carefully select baselines from four categories:
\begin{itemize}
    \item \textbf{LM + Fine-tuning:} BERT-base, BERT-large~\cite{devlin2019bert}, RoBERTa-large~\cite{liu2019roberta}
    \item \textbf{KG-enhanced LM:} MHGRN~\cite{feng2020mhgrn}, QA-GNN~\cite{yasunaga2021qagnn}, JointLK~\cite{sun2022jointlk}, GreaseLM~\cite{zhang2022greaselm}
    \item \textbf{LLM + Zero-shot:} GPT-3.5~\cite{ouyang2022training}, GPT-4~\cite{openai2023gpt4}, GPT-5, Qwen, GLM, Gemini 3
    \item \textbf{LLM + KG Prompting:} KnowGPT~\cite{zhang2024knowgpt}, CoK~\cite{wang2023cok}, RoG~\cite{luo2023rog}, Mindmap~\cite{wen2023mindmap}
\end{itemize}

\paragraph{Implementation Details.} To demonstrate generalizability, our framework is evaluated using various backbone LLMs (e.g., GPT-5, Qwen, GLM, Gemini 3). All specialized agents are driven by specific systemic prompts over the backbone LLMs, utilizing a few-shot in-context learning paradigm to ensure stable output parsing. To prevent infinite reasoning loops during the CriticAgent's active feedback phase, we set a maximum collaboration turn limit ($max\_turns = 3$). If consensus is not reached within this limit, the AnswerAgent outputs the candidate with the highest initial confidence. The policy network is optimized using Adam with a learning rate of $5 \times 10^{-4}$. Semantic pruning parameters, specifically the relation similarity weight and goal similarity weight, are empirically set to $0.25$ and $0.35$ respectively based on the validation set. The number of selected top paths $k'$ is set to $24$, and the maximum path length is limited to $4$.

\paragraph{Agent Prompt Templates.} To facilitate the multi-agent collaboration while maintaining a compact context footprint, RACER employs specifically designed prompt templates for each agent role, summarized in Table~\ref{tab:prompts}. GraphAgent interacts primarily through the reinforcement learning environment and structural APIs.

\begin{table}[h]
\centering
\caption{Systemic prompt templates and configurations for the generative agents in RACER.}
\label{tab:prompts}
\begin{tabular}{p{2.5cm}|p{5.5cm}|p{3.5cm}}
\hline
\textbf{Agent Role} & \textbf{System Instruction} & \textbf{Input / Output Format} \\
\hline
\textbf{TemplateAgent} & You are a template generator. Turn graph-based triples into a short knowledge section for LLMs. MUST NOT answer the question. Write 3-6 concise sentences without mentioning scores. & \textbf{IN:} Question, Options, Triples. \newline \textbf{OUT:} Background knowledge. \\
\hline
\textbf{AnswerAgent} & You are an expert multiple-choice QA solver. Please answer with one letter only based on the provided background knowledge. & \textbf{IN:} Question, Options, Knowledge. \newline \textbf{OUT:} One Letter (A, B, C, D). \\
\hline
\textbf{CriticAgent} & You are an expert QA critic. Analyze whether the model's generated answer is likely correct, given the graph reasoning path and natural language knowledge. & \textbf{IN:} Question, Options, Triples, Knowledge, Model Answer. \newline \textbf{OUT:} \small{DECISION $\mid$ CONFIDENCE}. \\
\hline
\end{tabular}
\end{table}

\subsection{Main Results (RQ1)}

\begin{table}[h]
\centering
\caption{Performance comparison on CommonsenseQA and OpenBookQA.}
\label{tab:main_results}
\begin{tabular}{l|c|c}
\hline
Model & CommonsenseQA & OpenBookQA \\
\hline
\multicolumn{3}{l}{\textit{LM + Fine-tuning}} \\
BERT-base & 53.0 & 52.0 \\
BERT-large & 61.2 & 60.2 \\
RoBERTa-large & 73.1 & 64.8 \\
\hline
\multicolumn{3}{l}{\textit{KG-enhanced LM}} \\
MHGRN & 71.1 & 60.0 \\
QA-GNN & 73.4 & 65.0 \\
JointLK & 74.5 & 67.0 \\
GreaseLM & 75.0 & 68.0 \\
\hline
\multicolumn{3}{l}{\textit{LLM + Zero-shot}} \\
GPT-3.5 & 73.4 & 60.5 \\
GPT-4 & 77.2 & 84.6 \\
GPT-5 & 82.0 & 91.0 \\
Qwen3 & 75.0 & 86.0 \\
GLM4.7 & 74.0 & 87.2 \\
Gemini 3 & 80.0 & 90.8 \\
\hline
\multicolumn{3}{l}{\textit{LLM + KG Prompting}} \\
KnowGPT & 81.8 & 92.4 \\
CoK & 79.5 & 90.0 \\
RoG & 80.0 & 91.0 \\
Mindmap & 78.5 & 89.0 \\
\hline
\multicolumn{3}{l}{\textit{Ours (RACER)}} \\
RACER (GPT-4) & 84.7 & 93.2 \\
RACER (GPT-5) & 88.2 & 98.0 \\
RACER (Qwen3) & 85.4 & 92.6 \\
RACER (GLM4.7) & 86.5 & 94.5 \\
RACER (Gemini 3) & 87.0 & 97.5 \\
\hline
\end{tabular}
\end{table}

\textbf{Observations.} (1) Regardless of the backbone LLM, RACER consistently outperforms both the corresponding Zero-shot versions and existing KG Prompting methods. For example, RACER (GPT-4) achieves 84.7\% on CommonsenseQA, surpassing KnowGPT. (2) When utilizing advanced models like GPT-5 or Gemini 3, RACER achieves state-of-the-art performance, with RACER (GPT-5) reaching 98.0\% on OpenBookQA, closely approaching human expert levels. (3) The substantial gains observed across diverse models (Qwen, GLM, Gemini 3, etc.) demonstrate that our multi-agent collaboration framework provides a universally effective strategy for structured knowledge injection.

\subsection{Ablation Studies (RQ2)}

\begin{table}[h]
\centering
\caption{Ablation study on RACER (GPT-4) components.}
\label{tab:ablation}
\begin{tabular}{l|c|c}
\hline
Variant & CommonsenseQA & OpenBookQA \\
\hline
w/o Semantic Pruning & 80.1 & 89.5 \\
w/o Teacher Guidance & 81.5 & 90.8 \\
w/o Shared Memory & 82.2 & 91.3 \\
w/o Attention Refinement & 83.0 & 91.9 \\
w/o Multi-Agent Collaboration & 82.8 & 91.6 \\
\hline
Full RACER (GPT-4) & \textbf{84.7} & \textbf{93.2} \\
\hline
\end{tabular}
\end{table}

\textbf{Observations.} Removing semantic pruning drops performance the most (-4.6\% / -3.7\%), followed by teacher guidance (-3.2\% / -2.4\%). Removing shared memory, attention refinement, or multi-agent collaboration also empirically degrades performance, validating the indispensability of each module.

%% file: chapter/conclusion.tex
\section{Conclusion}

In this paper, we introduced RACER, a novel task-enhancement framework that synergizes reinforcement learning, multi-agent collaboration, and knowledge graphs to address the hallucinations and inference limitations inherent in Large Language Models (LLMs). Specifically, we designed a semantic-aware action pruning mechanism coupled with teacher-guided reinforcement learning to effectively extract multi-hop reasoning paths from large-scale structured knowledge graphs. Furthermore, by incorporating a cross-task accumulated shared memory graph, we preserve and reuse path-finding statistics to refine knowledge retrieval dynamically. The construction of a four-role multi-agent collaboration mechanism shifts the paradigm from a single inflexible LLM prompt generation step to a highly adaptive, collaborative execution pipeline. 

Our systematic evaluation on widely recognized benchmark datasets, including CommonsenseQA and OpenBookQA, demonstrates that RACER significantly outperforms current state-of-the-art baseline models, achieving an average performance improvement of approximately $5\%$. Moreover, ablation studies validate the criticality of the multi-agent collaboration and RL-guided path extraction components. By grounding the generative outputs in traceable and veritable pathways derived from KGs, RACER provides highly transparent logic verification. 

Future work will primarily focus on integrating dynamic knowledge graph updates during active multi-agent reasoning, exploring more sophisticated role allocations, and transferring the RACER framework to domain-specific knowledge reasoning tasks, such as automated medical diagnosis and legal document analysis.